\documentclass{article}

\usepackage{microtype}
\usepackage{graphicx}
\usepackage{subfigure}
\usepackage{booktabs} % for professional tables

\usepackage{hyperref}

\usepackage[preprint]{icml2025}

\usepackage{amsmath}
\usepackage{amssymb}
\usepackage{mathtools}
\usepackage{amsthm}

\usepackage{hyperref}       % hyperlinks
\usepackage{url}            % simple URL typesetting
\usepackage{booktabs}       % professional-quality tables
\usepackage{amsfonts}       % blackboard math symbols
\usepackage{nicefrac}       % compact symbols for 1/2, etc.
\usepackage{microtype}      % microtypography
\usepackage{xcolor}         % colors
\usepackage{amsmath}
\usepackage{graphicx}
\usepackage{wrapfig}
\usepackage{adjustbox}
\usepackage{booktabs}
\usepackage{multirow}
\usepackage{enumitem} 
\usepackage{array}
\usepackage{caption}
\usepackage{hyperref} 
\usepackage{makecell}
\usepackage{tcolorbox}
\usepackage{subcaption}
\usepackage{listings}
\usepackage{graphicx}
\usepackage{wrapfig,lipsum,booktabs}
\usepackage{natbib}
\usepackage{fontawesome5} 
\usepackage{colortbl}
\definecolor{PsuBlue}{HTML}{1E407C}

\newcommand{\SampleBox}[3][]{%
  \begin{tcolorbox}[
    colframe=PsuBlue,      
    colback=PsuBlue!5!white,  
    title={#2}
  ]
    \IfFileExists{#3}{%
      \lstinputlisting[style=sampletxt,#1]{#3}%
    }{%
      \textit{File not found: \texttt{#3}. Check path/case in Overleaf.}%
    }%
  \end{tcolorbox}%
}

\lstdefinestyle{sampletxt}{
  basicstyle=\footnotesize\ttfamily,   % Use \small\rmfamily if you prefer body text
  breaklines=true,
  breakatwhitespace=false,
  postbreak=\mbox{\textcolor{gray}{$\hookrightarrow$}\space},
  columns=fullflexible,
  keepspaces=true,
  showstringspaces=false,
  tabsize=2,
  literate=
    {_}{{\_}}1
    {\%}{{\%}}1
    {\$}{{\$}}1
    {&}{{\&}}1
}

\usepackage[capitalize,noabbrev]{cleveref}

\theoremstyle{plain}

\theoremstyle{definition}

\theoremstyle{remark}

\usepackage[textsize=tiny]{todonotes}
\icmltitlerunning{Simple Denoising Diffusion Language Models}

\begin{document}

\twocolumn[
\icmltitle{Simple Denoising Diffusion Language Models}

% It is OKAY to include author information, even for blind
% submissions: the style file will automatically remove it for you
% unless you've provided the [accepted] option to the icml2025
% package.

% List of affiliations: The first argument should be a (short)
% identifier you will use later to specify author affiliations
% Academic affiliations should list Department, University, City, Region, Country
% Industry affiliations should list Company, City, Region, Country

% You can specify symbols, otherwise they are numbered in order.
% Ideally, you should not use this facility. Affiliations will be numbered
% in order of appearance and this is the preferred way.
% \icmlsetsymbol{equal}{*}

\begin{icmlauthorlist}
\icmlauthor{Huaisheng Zhu}{yyy}
\icmlauthor{Zhengyu Chen}{comp}
\icmlauthor{Shijie Zhou}{sch}
\icmlauthor{Zhihui Xie}{hks}
\icmlauthor{Yige Yuan}{comp2}
\icmlauthor{Shiqi Chen}{hks2}
\icmlauthor{Zhimeng Guo}{yyy}
\icmlauthor{Siyuan Xu}{yyy}
\icmlauthor{Hangfan Zhang}{yyy}
%\icmlauthor{}{sch}
\icmlauthor{Vasant Honavar}{sch}
\icmlauthor{Teng Xiao}{sch2,comp3}
%\icmlauthor{}{sch}
%\icmlauthor{}{sch}
\end{icmlauthorlist}

% \icmlaffiliation{yyy}{Department of XXX, University of YYY, Location, Country}
 \icmlaffiliation{yyy}{Penn State University}
\icmlaffiliation{comp}{Meituan}
\icmlaffiliation{sch}{University at Buffalo}
\icmlaffiliation{sch2}{University of Washington}
\icmlaffiliation{hks}{The University of Hong Kong}
\icmlaffiliation{hks2}{City University of Hong Kong}
\icmlaffiliation{comp2}{Alibaba Group}
\icmlaffiliation{comp3}{Allen Institute for AI (AI2)}

\icmlcorrespondingauthor{Huaisheng Zhu}{hvz5312@psu.edu}
% \icmlcorrespondingauthor{Firstname2 Lastname2}{first2.last2@www.uk}

% You may provide any keywords that you
% find helpful for describing your paper; these are used to populate
% the "keywords" metadata in the PDF but will not be shown in the document
\icmlkeywords{Machine Learning, ICML}

\vskip 0.3in
]

% this must go after the closing bracket ] following \twocolumn[ ...

% This command actually creates the footnote in the first column
% listing the affiliations and the copyright notice.
% The command takes one argument, which is text to display at the start of the footnote.
% The \icmlEqualContribution command is standard text for equal contribution.
% Remove it (just {}) if you do not need this facility.

\printAffiliationsAndNotice{}  % leave blank if no need to mention equal contribution
% \printAffiliationsAndNotice{\icmlEqualContribution} % otherwise use the standard text.

\begin{abstract}

Recent Uniform State Diffusion Models (USDMs), initialized from a uniform prior, offer the promise of fast text generation due to their inherent self-correction ability compared to masked diffusion models. However, they still rely on complex loss formulations with additional computational overhead, which hinders scalability. In this work, we explore a simplified denoising-based loss for USDMs that optimizes only noise-replaced tokens, stabilizing training while matching the performance of prior methods with more complex objectives. In addition, we introduce an efficient regularization term to mitigate corruption toward uniform output distributions, which further improves performance. We demonstrate the effectiveness and efficiency of our simple and improved loss formulations by pretraining models on widely used text datasets for USDMs. More importantly, our conclusions scale to larger models, showing strong potential for large-scale training. The code of our method is available at this \href{https://github.com/huaishengzhu/Simple-Denoising-Diffusion-Language-Models}{link}.
\end{abstract}

\section{Introduction}

Diffusion models are powerful generative frameworks that excel at producing realistic, high-quality continuous data such as images and videos~\cite{ho2020denoising, song2020denoising, rombach2022high, kong2020diffwave, ho2022video}. They achieve this by training denoising models to reconstruct samples corrupted with varying levels of Gaussian noise. Generation then proceeds through a Markov chain: starting from pure noise, the model iteratively denoises the sample, gradually transforming it into a clean image.

To further advance the capabilities of diffusion models, Masked Diffusion Models (MDMs), which use the prior distribution by masking all tokens, have recently demonstrated remarkable progress across language generation tasks~\cite{sohl2015deep,austin2021structured,campbell2022continuous,lou2023discrete,meng2022concrete}. By optimizing the simpilfied varaint of evidence lower bound (ELBO), masked diffusion language models have achieved performance comparable to, and in some cases surpassing, that of autoregressive models (ARMs)~\cite{sahoo2024simple, shi2024simplified, nie2025large}. Moreover, recent studies have investigated the scaling properties of MDMs, demonstrating that they can achieve competitive performance with advanced autoregressive models of similar size (e.g., Llama 2~\cite{touvron2023llama} and Llama 3~\cite{dubey2024llama}) on a range of downstream tasks~\cite{nie2025large, gong2024scaling, nie2024scaling, gong2025diffucoder, ye2025dream}.

Despite its great success, MDMs experience severe performance degradation in the few-step regime~\cite{deschenaux2024beyond}. In contrast to diffusion models with Probability Flow ODEs in continuous space~\cite{song2020score}, MDMs lack an implicit property—a deterministic mapping from noise to data. To address this limitation, recent work on Uniform State Diffusion Models (USDMs) explore language modeling by initializing from a uniform distribution, analogous to Gaussian noise in continuous diffusion models~\cite{sahoo2025diffusion, austin2021structured, zhao2024informed, schiff2024simple}. Inspired by the extensive success of Gaussian diffusion, these models benefit from a range of well-studied techniques, especially efficient training and  distillation schemes that enable fast generation~\cite{song2023consistency}. These models achieve performance comparable to MDMs while demonstrating strong potential to reduce the number of sampling steps without compromising generation quality. Moreover, USDMs exhibit superior scaling behavior compared to MDMs as training FLOPs increase, demonstrating their promise as a direction for diffusion-based language models~\cite{von2025scaling}.

\begin{figure*}[t]
    \centering
    \includegraphics[width=0.99\linewidth]{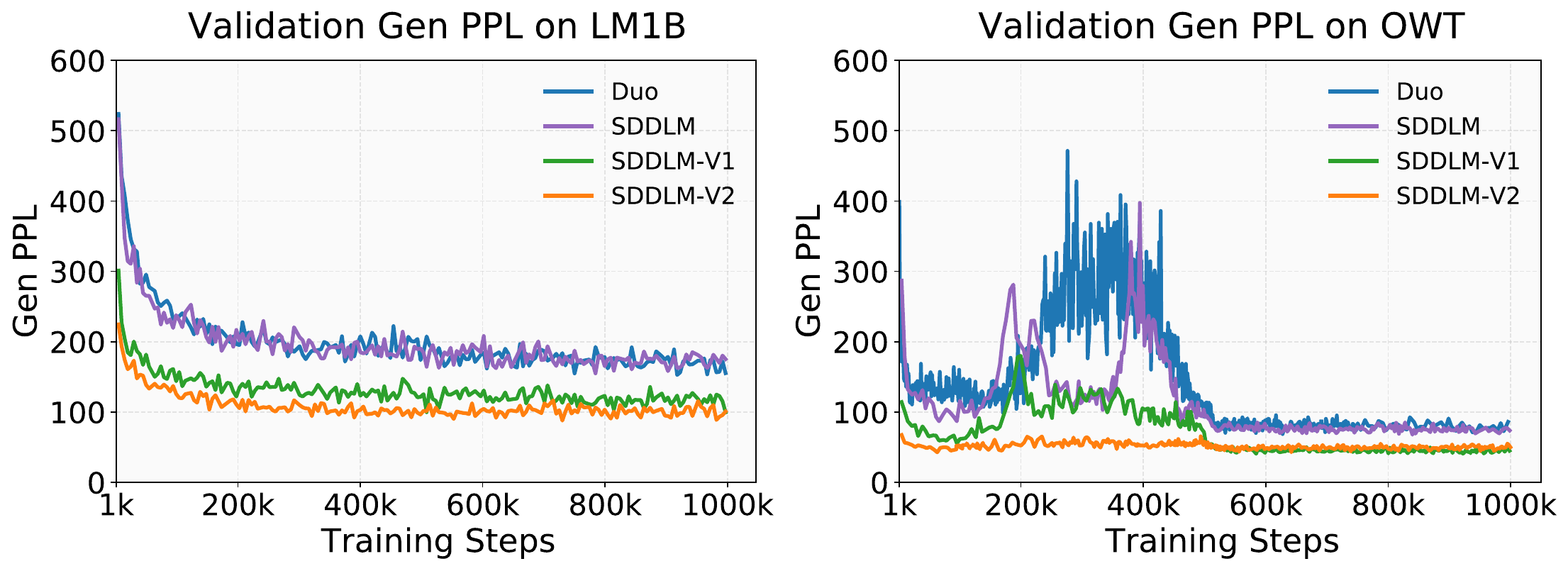}
    \vspace{-0.9em}
    \caption{Validation Gen PPL of different models over training steps.}
    \label{fig:ppl_curve_lm1b}
\end{figure*}
However, the current state-of-the-art USDMs~\cite{sahoo2025diffusion}, which adopts uniform distributions as the prior, still suffers from a complex loss formulation. This complexity leads to additional computational overhead during training and may hinder scalability. Therefore, in this paper, it naturally inspires us to \textit{explore how to simplify the training objective for USDMs with great scaling abilities}.
% \textit{How can we design a simple, efficient and effective algorithm for Uniform-state Diffusion Models?}

To resolve this problem, we explore a simpler loss formulation for diffusion language models initialized from a uniform prior (i.e., pure noise). Specifically, analogous to diffusion models with Gaussian noise in continuous spaces~\cite{ho2020denoising}, we start from standard denoising objectives, in which noise is added to clean sequences and the model is trained to denoise the corrupted inputs back to the original sequences. However, we find that this naive adaptation frequently leads to training collapse. Motivated by the design of MDMs, we instead propose a method called Simple Denoising Diffusion Language Model (SDDLM),  which optimizes only the tokens that are replaced with noise. This strategy, akin to the selective denoising behavior of MDMs, stabilizes training and achieves performance comparable to the ELBO-derived loss, while avoiding its significant computational cost. Moreover, we introduce Anti-Uniform Distribution Sharpening regularization to mitigate prediction corruption caused by noisy sequences that tend to induce near-uniform output distributions. From a self-supervised learning perspective, the resulting negative gradient is closely related to contrastive learning~\cite{xiang2023denoising, chen2024deconstructing}. As shown in Figure~\ref{fig:entropy_curve_lm1b}, our results with SDDLM-V1 demonstrate that this design yields notable improvements in generation quality.
% we further interpret this denoising process as a form of self-supervised learning, a perspective that has also been explored in diffusion models on continuous spaces~\cite{xiang2023denoising, chen2024deconstructing}. Building on this perspective, we incorporate negative gradients, motivated by other effective self-supervised learning methods such as contrastive learning, into the training process. As shown in Figure~\ref{fig:entropy_curve_lm1b}, our results demonstrate that this design leads to notable improvements in generation quality.

\noindent\textbf{Contributions.} 
% \noindent\textbf{Main Contributions.}
The main contributions of this paper are as follows:
(i) We introduce a \emph{simple and efficient denoising-based training framework} for USDMs that substantially simplifies prior training objectives while retaining strong empirical performance.
(ii) We further improve performance by proposing a \emph{regularization term with negative gradients} (SDDLM-V1), motivated by a self-supervised and contrastive learning interpretation of the denoising objective, which mitigates degeneration toward uniform output distributions.
(iii) We demonstrate that SDDLM achieves \emph{competitive performance} relative to prior ELBO-based approaches that incur higher computational overhead during training. Also, SDDLM-V1 consistently improves generation quality. Importantly, these results \emph{scale to larger models (up to 1.1B parameters)}, indicating indicating strong effectiveness for large-scale pretraining.

%The main contributions of this paper are: (i) To the best of our knowledge, we are the first to explore a simple and efficient framework that leverages a denoising objective to simplify previous training methods of USDMs. 
% We propose the Simple Denoising Diffusion Language Model (SDDLM), 
%(ii) We further improve performance by introducing a regularization term with negative gradients (denoted as SDDLM-V1), which is analogous to contrastive learning by interpreting denoising objective from self-supervised learning perspectives.
% reinterpret this denoising objective from a self-supervised learning perspective and introduce a contrastive learning–based loss that provides negative gradients, resulting in SDDLM-V1, which further improves model performance. 
%(iii) SDDLM achieves competitive performance compared to prior methods that optimize the ELBO, which incur additional computational overhead during training. Moreover, our SDDLM-V1 further improves generation performance through a negative gradient formulation. More importantly, our method scales effectively to larger models (1.1B) and yields consistent conclusions, demonstrating strong potential for future large-scale pretraining.

\section{Background}

\subsection{Denoising Diffusion Models}\label{background:gaussian}
Denoising diffusion models on continuous space formulate generation as a Markov process that transforms the data distribution $q_\text{data}$ into a simple prior on continuous space, such as a standard normal distribution $\mathcal{N}(0, \mathbf{I})$. Concretely, the process begins with samples from the data distribution and iteratively adds noise to produce a sequence of noisy latents $\mathbf{x}_t \sim q_t(\cdot \mid \mathbf{x}_0)$, whose marginal distribution is:
\begin{align}\label{eqn:gaussian_marginal}
    q_t\left(\cdot \mid \mathbf{x}_0 ; \bar{\alpha}_t\right)=\mathcal{N}\left(\mathbf{x}_t ; \sqrt{\bar{\alpha}_t} \mathbf{x}_0,\left(1-\bar{\alpha}_t\right) \boldsymbol{I}\right),
\end{align} 
where the diffusion parameter $\bar{\alpha}_t \in [0, 1]$ is a monotonically decreasing function in $t$ and $\mathbf{x}_0 \sim q_{\text{data}}$. Then, a simplified version of evidence lower bound (ELBO) based on denosing noisiy images into clean images is minimized to train the diffusion model with the following equation:
\begin{align}\label{eqn:gaussian_elbo}
    \mathbb{E}_{\mathbf{x}_0, t, \mathbf{\epsilon}}\left[\lambda(t)\left\|\mathbf{x}_0-\mathbf{x}_\theta\left(\mathbf{x}_t, t\right)\right\|^2\right]
\end{align}
where $\boldsymbol{\epsilon} \sim \mathcal{N}(0, \boldsymbol{I}), t \sim \mathcal{U}(0, T), \mathbf{x}_t \sim q_t\left(\cdot \mid \mathbf{x}, \bar{\alpha}_t\right)$. $\lambda(t)$ is a time dependent weighting function and can be ignored during the training process. $\theta$ are learnable parameters.

\subsection{Discrete Diffusion Models}
Previous objectives and formulations are primarily based on Gaussian distributions and operate in continuous spaces. To adapt diffusion models to discrete data $\mathbf{x} \in \mathcal{V}$, where $\mathcal{V}$ denotes the vocabulary for language generation, the discrete diffusion framework~\citep{sohl2015deep,austin2021structured} extends the core idea of continuous denoising diffusion models: mapping the data distribution $q_\text{data}$ to a simple prior distribution through a sequence of Markov states. Similar to their continuous counterparts, the noise-adding process—referred to as the forward process $(q_t)(t \in [0,1])$—smoothly transitions from $q_\text{data}$ to a categorical prior $\operatorname{Cat}(\cdot; \boldsymbol{\pi})$ by interpolating between the data distribution and the prior. The corresponding marginals conditioned on one token $\mathbf{x}_0^l$ at time $t$ are given by the following equation:
\begin{align}\label{eqn:discrete_marginal}
    q_t\left(. \mid \mathbf{x}^l_0 ; \alpha_t\right)=\operatorname{Cat}\left(. ; \alpha_t \mathbf{x}^l_0+\left(1-\alpha_t\right) \boldsymbol{\pi}\right).
\end{align}
For MDLMs, the prior $\boldsymbol{\pi}$ is typically defined using a special masked token, i.e., $\boldsymbol{\pi} = \mathbf{M}$ with $\mathbf{M} \in \mathcal{V}$~\citep{sahoo2024simple}. Alternatively, a uniform prior can be defined as $\boldsymbol{\pi} = \mathbf{1}/V$, where $V = |\mathcal{V}|$ denotes the vocabulary size. Specifically, in MDMs, a token $\mathbf{x}$ either stays unchanged or is replaced by the mask token $\mathbf{m}$, remaining masked thereafter. In USDMs, each token can instead transition uniformly to any token in $\mathcal{V}$, with probabilities determined by the diffusion timestep. To train USDMs, the Negative Evidence Lower Bound (NELBO) loss for the token $l$ is derived using principles similar to those of continuous diffusion models, and can be expressed in the following form~\citep{lou2023discrete, schiff2024simple, sahoo2025diffusion}:
\begin{equation}
\label{eq:usdm}
\begin{aligned}
     & \mathcal{L}^l_{\text{USDM}} = \mathbb{E}_{t \sim \mathcal{U}[0,1], q_t\left(\mathbf{x}^l_t \mid \mathbf{x}^l_0 ; \alpha_t\right)}-\frac{\alpha_t^{\prime}}{V \alpha_t}\left[\frac{V}{\tilde{\mathbf{x}}^l_i} \right. \\
     & \left. -\frac{V}{\left(\tilde{\mathbf{x}}^l_\theta\right)_i}  -\sum_j \frac{\tilde{\mathbf{x}}^l_j}{\tilde{\mathbf{x}}^l_i} \log \frac{\left(\tilde{\mathbf{x}}^l_\theta\right)_i \cdot \tilde{\mathbf{x}}^l_j}{\left(\tilde{\mathbf{x}}^l_\theta\right)_j \cdot \tilde{\mathbf{x}}^l_i}\right],
\end{aligned}
\end{equation}
where $\tilde{\mathbf{x}}^l=V \alpha_t \mathbf{x}_0^l+\left(1-\alpha_t\right) \mathbf{1}$, $\tilde{\mathbf{x}}^l_\theta=V \alpha_t \mathbf{x}^l_\theta\left(\mathbf{x}_t, t\right)+\left(1-\alpha_t\right) \mathbf{1}$, $\mathbf{x}^l_t \sim q_t(\cdot \mid \mathbf{x}^l_0;\alpha_t)$ and $\alpha_t^{\prime}$ is the time-derivative of the $\alpha_t$. $i=\arg \max _{j \in[V]}\left(\mathbf{x}^l_t\right)_j$ is the non-zero entry of $\mathbf{x}^l_t$. Other studies have also explored more efficient approaches to computing the loss in Equation~(\ref{eq:usdm})~\cite{sahoo2025diffusion}. $\mathbf{x}^l_\theta$ denotes a neural network $\mathcal{V} \times[0,1] \rightarrow \Delta^V$, where $\Delta^V$ denotes the K-simplex and $\theta$ are trainable parameters . After training models with Equation (\ref{eq:usdm}), USDMs typically generate samples by applying the reverse diffusion process, starting from the uniform prior in the following equation:
\begin{equation}
\label{eq:reverse}
    \begin{aligned}
&q_{s \mid t}\left(. \mid \mathbf{x}^l_t, \mathbf{x}^l_0\right) =\operatorname{Cat}\left(; \frac{V \alpha_t \mathbf{x}^l_t \odot \mathbf{x}^l_0+\left(\alpha_{t \mid s}-\alpha_t\right) \mathbf{x}^l_t}{V \alpha_t\left\langle\mathbf{x}^l_t, \mathbf{x}^l\right\rangle+1-\alpha_t}\right.\\
 & +  \left.\frac{\left(\alpha_s-\alpha_t\right) \mathbf{x}^l_0+\left(1-\alpha_{t \mid s}\right)\left(1-\alpha_s\right) \mathbf{1} / V}{V \alpha_t\left\langle\mathbf{x}^l_t, \mathbf{x}^l_0\right\rangle+1-\alpha_t}\right),
\end{aligned}
\end{equation}
where $s<t$, $\alpha_{t \mid s}=\alpha_t / \alpha_s$ and $\mathbf{x}^l$ represents a one-hot vector, with a nonzero entry indicating the token. During inference, we replace $\mathbf{x}$ in Equation (\ref{eq:reverse}) with $\mathbf{x}_\theta (\mathbf{x}_t, t)$. In the following section, we use $p_\theta(\mathbf{x}_0 \mid \mathbf{x}_t)$ to denote $\mathbf{x}_\theta (\mathbf{x}_t, t)$.

\section{Related Works}

\textbf{Denoising Diffusion Models.} Denoising diffusion probabilistic models have proven to be powerful tools for generating diverse data types~\cite{ho2020denoising, song2020denoising}. The sampling process of
diffusion models can be interpreted as stochastic differential equations (SDEs) and is trained using
score matching objectives based on this formulation~\cite{song2020score}. Moreover, the strong generative performance of Denoising Diffusion Models (DDMs) has attracted increasing interest in their potential for representation learning~\cite{xiang2023denoising, chen2024deconstructing}, as their training process is equivalent to that of Denoising Autoencoders (DAEs)~\cite{vincent2008extracting}, which remove noise at multiple levels through a diffusion-driven procedure. Interestingly, aligning denoising diffusion models with self-supervised learning–based models enables efficient training of generative diffusion models~\cite{yu2024representation}. This insight motivates us to reconsider the training objective for USDMs from a self-supervised learning perspective, where a denoising objective enables diffusion models to be trained in a continuous space. Moreover, contrastive learning~\cite{chen2020simple}, by providing informative negative gradients, can further enhance representation learning, which in turn inspires us to further explore training objectives for USDMs.

\noindent \textbf{Diffusion Lanuage Models.} The development of DLLMs is motivated by recent advances in discrete diffusion models, which introduced new forward and reverse transition mechanisms and enabled a diverse range of model variants~\cite{sohl2015deep,austin2021structured,campbell2022continuous,lou2023discrete,meng2022concrete}. Empirical studies further demonstrate that masked diffusion models (MDMs) can achieve perplexity comparable to autoregressive models (ARMs)~\cite{sahoo2024simple,shi2024simplified,nie2025large,ou2024your}. To improve training efficiency, several works have proposed simplified training objectives for masked diffusion processes with theoretical justifications. In addition, recent research has examined the scaling behavior of MDMs, including both training from scratch and adaptation from pre-trained ARMs~\cite{nie2025large, gong2024scaling, nie2024scaling, ni2025training, nidiffusion}.  Although MDLMs demonstrate greater efficiency than ARMs by generating multiple tokens simultaneously, they suffer from notable performance degradation in the few-step generation regime~\cite{deschenaux2024beyond}. While numerous techniques for reducing sampling steps without sacrificing generation quality have been successful in continuous-space diffusion models, directly transferring these methods to MDMs is difficult, as they lack the inherent property of mapping noise to data. To overcome this limitation, recent works on Uniform-state Diffusion Models (USDMs) have explored initializing language models from a uniform distribution, analogous to Gaussian noise in continuous diffusion~\cite{sahoo2025diffusion, austin2021structured, zhao2024informed, schiff2024simple}. Unlike MDLMs, which simplify the training loss and scale effectively to larger models, current USDMs still rely on complex ELBO-derived losses, potentially limiting their scalability. Therefore, in this paper, we study the problem of simplifying the loss of USDMs.

\section{Method}
In this section, we investigate the use of a simple denoising loss to improve USDMs. We first observe that a naive application of this loss can lead to performance degradation. To address this issue, we apply the denoising loss only to perturbed positions, which effectively mitigates the observed degradation, denoted as Simple Denoising Diffusion Language Model (SDDLM). Building on this, we further introduce an anti-uniform distribution sharpening regularization that incorporates negative gradients and denote this method as SDDLM-V1. This regularization counteracts the tendency toward uniform distribution corruption during training, leading to more stable optimization and improved final performance. Finally, we connect our design with contrastive learning and inspire other design in the future work.
\begin{figure}
    \centering
    \includegraphics[width=\linewidth]{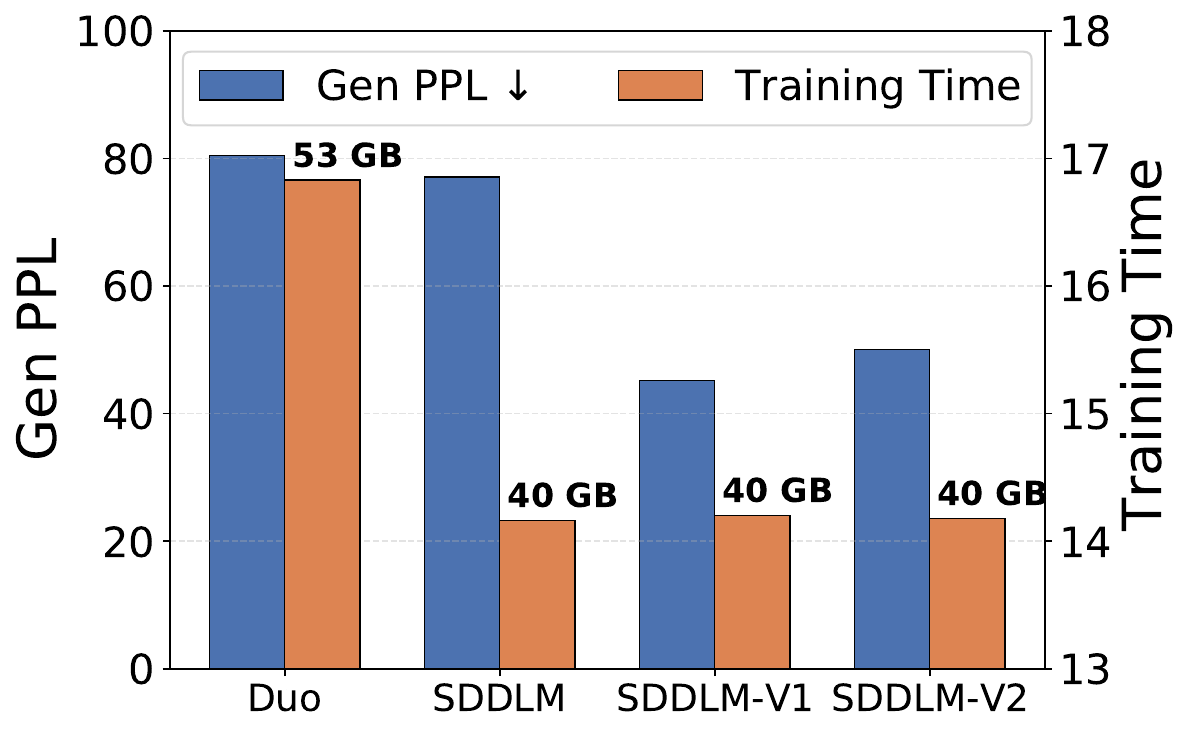}
    \caption{Running time, memory and generation quality comparison on OWT. We report the training time for 0.5 epochs and the peak GPU memory for each method under the same batch size, together with their generation quality.}
    \label{fig:time}
\end{figure}
\subsection{Simple Denoising Diffusion Language Models}
Comparing the training loss of continuous diffusion models in Equation~(\ref{eqn:gaussian_elbo}) with that of USDMs in Equation~(\ref{eq:usdm}), we observe that the latter is considerably more complex. To address this, we propose a simplified formulation of the USDMs loss objective.  First, we consider optimizing an objective analogous to the reconstruction loss in Equation~(\ref{eqn:gaussian_elbo}), where the model directly learns to predict the original sequence (the desired target) by replacing the clean sequence $\mathbf{x}_0$ in Equation~(\ref{eq:reverse}). This approach is also used in training discrete flow-matching models~\cite{gat2024discrete}:
\begin{equation}
\label{eq:naive}
    \min_\theta \mathbb{E}_{\mathbf{x}_0, t, \mathbf{x}_t \sim \mathcal{D},  \mathcal{U}[0,1], q_t} \sum_{l=1}^L -\log p_\theta (\mathbf{x}_0^l \mid \mathbf{x}_t),
\end{equation}
where $\mathbf{x}_0$ are sequences of tokens from $\mathcal{D}$ and $\mathbf{x}_t$ is used to reconstruct $\mathbf{x}_0$ with a denoising loss and each token of $\mathbf{x}$ is perturbed by Equation (\ref{eqn:discrete_marginal}). However, training with this loss leads to degraded model performance and unstable optimization. Note that we exclude this baseline from comparisons due to unstable convergence.
% However, this naive adour empirical studies of sampling with the reverse process in Equation~(\ref{eq:reverse}) show that relying on a reconstruction-based loss objective degrades model performance. 
We attribute this issue to the structure of the sequence $\mathbf{x}_t$, which consists of two parts: (i) positions where $\mathbf{x}_t^j \neq \mathbf{x}_0$, corresponding to tokens corrupted by noise, and (ii) positions where $\mathbf{x}_t^j = \mathbf{x}_0$, which remain unchanged. When applying a reconstruction-based objective, these two parts impose different learning goals: the noisy positions require denoising, while the unchanged positions reduce to reconstructing the input itself. During our empirical studies, we find that the denoising component is more critical, as the model must reconstruct clean sequences through its denoising ability. This process is also closely aligned with the training paradigm of MDMs, which predict only the masked tokens; this perspective also inspires us to focus prediction on the noisy components. Guided by this observation, we propose focusing the objective on the denoising part of the sequence by the following equation:
\begin{equation}
\begin{aligned}
    \label{eq:denoising}
     & \mathcal{L}_\text{SDDLM} = \\
      & \mathbb{E}_{\mathbf{x}_0, t, \mathbf{x}_t \sim \mathcal{D},  \mathcal{U}, q_t} \sum_{l=1}^L -\log p_\theta (\mathbf{x}_0^l \mid \mathbf{x}_t) \mathbf{1}\left[\mathbf{x}_0^l\neq\mathbf{x}^l_t\right].
\end{aligned}
\end{equation}
This simple objective achieves generation performance comparable to the original, more complex NEBLO loss in Equation (\ref{eq:usdm}), as shown in Figure~\ref{fig:ppl_curve_lm1b}. Moreover, it incurs substantially lower training cost while maintaining comparable performance, as illustrated in Figure~\ref{fig:time}. The training details of algorithms are put into Appendix~\ref{sec:imple}.

\begin{table*}[t!]
    \centering
    \vskip -0.7em
    \begin{adjustbox}{max width=\textwidth}
    \begin{tabular}{lccccccccc}
      \toprule
         &  BoolQ & Hellaswag &  Obqa &  PIQA & RACE & SIQA  & LAMBADA \\
        \midrule
        % Duo & 42.85 & \textbf{60.55} & 34.47 & 17.00 & 63.11 & 31.39 & 37.82 & 17.02 \\
        SDDLM & 60.34 & 34.60 & 18.20 & 64.42 & 30.05 & 37.56 & 26.53 \\
        SDDLM-CFG & \textbf{62.11} & \textbf{36.09} & \textbf{20.00} & \textbf{64.80} & \textbf{30.72} & \textbf{37.87} & \textbf{32.60} \\
      \bottomrule
    \end{tabular}
    \end{adjustbox}
    % \vskip -0.7em
    \caption{Results for CFG methods on 1.1B models.}
    \label{tab:compare_large_cfg}
\end{table*}

\subsection{Anti-uniform Distribution Sharpening}
So far, we have focused on a simple denoising objective to improve training efficiency. We now explore whether this formulation can be further improved in terms of overall performance, while retaining its simplicity. USDMs employ a uniform prior over the vocabulary, which introduces strong stochasticity during the corruption process compared with AR models and MDMs.  This simple denoising loss induces a characteristic failure mode: under heavy corruption, the conditional distribution tends to become overly smooth and close to uniform. In such cases, a standard denoising objective that only rewards the ground-truth token may provide insufficient discriminative signal, leading to high-entropy predictions and degraded generation quality. To mitigate this effect, we introduce an anti-uniform distribution sharpening regularizer that explicitly encourages the model’s conditional prediction to deviate from the uniform distribution. Concretely, in addition to maximizing the likelihood of the ground-truth token, we penalize probability mass assigned to randomly sampled tokens from the vocabulary. The resulting objective is shown as follows:
\begin{equation}
\label{eq:plusregular}
\begin{aligned}
      \mathcal{L}_{\text{SDDLM}} - \sum_{l=1}^L \operatorname{KL}\left(\mathcal{U} \| p_\theta\left(\mathbf{x}_0^l \mid \mathbf{x}_t\right)\right)\mathbf{1}\left[\mathbf{x}_0^l\neq\mathbf{x}^l_t\right],
\end{aligned}
\end{equation}
where $\mathcal{U}$ denotes the uniform distribution over the vocabulary $\mathcal{V}$, i.e., $\mathcal{U} = 1/|\mathcal{V}|$ for all $v \in \mathcal{V}$. Thus, minimizing this objective implicitly maximizes the discrepancy between the uniform distribution and the model’s predictions, encouraging the model to move away from uniform outputs and produce sharper, more discriminative token distributions. This effect  counteracts the over-smoothing induced by uniform-state corruption. After simple mathematical derivation (in Appendix~\ref{sec:dervinegative}), we get the following equation:
% This perspective has also been explored in continuous diffusion models~\cite{chen2024deconstructing}. Beyond simple denoising losses or denoising autoencoders, contrastive learning is also widely used in self-supervised learning and relies on negative gradients. Therefore, we aim to further enhance the training process by incorporating negative gradients, inspired by Noise Contrastive Estimation (NCE) or contrastive learning. However, directly applying NCE differs significantly from the loss defined in Equation (\ref{eq:denoising}). In practice, we observe that this approach leads to unstable training and strong sensitivity to the number of negative samples, requiring careful tuning or selection of negative examples. To simplify this and make it practical for pretraining, we propose the following objective, which just randomly selects one token for contrastive comparison:
\begin{equation}
\begin{aligned}
    \label{eq:negative}
    & \mathcal{L}_\text{SDDLM-V1} = \\
    & \mathbb{E}_{\mathbf{x}_0, t, \mathbf{x}_t \sim \mathcal{D},  \mathcal{U}, q_t} \sum_{l=1}^L (-\log p_\theta (\mathbf{x}_0^l  \mid \mathbf{x}_t) \\ 
    & + \mathbb{E}_{\hat{\mathbf{x}}^l \sim \operatorname{U}(\mathcal{V})} \log p_\theta (\hat{\mathbf{x}}^l \mid \mathbf{x}_t))\mathbf{1}\left[\mathbf{x}_0^l\neq\mathbf{x}^l_t\right],
\end{aligned}
\end{equation}
where $\hat{\mathbf{x}}^l \sim \operatorname{U}(\mathcal{V})$ denotes a token randomly sampled from the vocabulary. However, during optimization, the negative gradient may dominate due to the large gradients induced by small values in the logarithmic terms, leading to training instability and model degradation. We add a small constant $\varepsilon$ to $p_\theta (\mathbf{x}_0^l  \mid \mathbf{x}_t)$ and $p_\theta (\hat{\mathbf{x}}_0^l  \mid \mathbf{x}_t)$ to stabilize the gradient of the logarithm. Note that the training of SDDLM without adding $\epsilon$ will directly lead to break for the model so we ignore this training ablation studies in our experiment section. Our proposed loss objective is a simple modification of Equation (\ref{eq:denoising}) and proves to be practical and effective.

\subsection{Connection with Contrastive Learning}

We propose a simple regularization term that substantially improves generation quality, as demonstrated by our empirical results. In this section, we further analyze and seek to understand the effects of this regularization term from other perspectives. Our first SDDLM adopts a denoising objective from a self-supervised learning perspective. We further interpret our improved version, SDDLM-V1 through the lens of contrastive learning~\cite{chen2020simple}, a prominent self-supervised paradigm, by analyzing its gradient structure. This structure comprises both attractive and repulsive gradient components, which are known to be effective for representation learning. From this perspective, SDDLM-V1 naturally incorporates both attractive and repulsive gradients, a property that may also be beneficial for leveraging negative gradients. In particular, the gradient of our loss (Equation (\ref{eq:negative})) takes the following form, which closely resembles that used in contrastive learning:
\begin{equation}
   - \nabla_\theta z_\theta\left(\mathbf{x}_0^l\right)+ \mathbb{E}_{\hat{\mathbf{x}}^l \sim \operatorname{U}(\mathcal{V})} \nabla_\theta z_\theta\left(\hat{\mathbf{x}}^l\right),
\end{equation}
where $z_\theta= \log p_\theta (\mathbf{x}_0^l\mid \mathbf{x}_t)$. This perspective from contrastive learning provides a useful interpretation of our method and suggests potential avenues for further improvement through self-supervised learning perspectives in future. Therefore, this insight further motivates SDDLM-V1 to design more principled negative sampling strategies beyond random sampling from a uniform distribution.

\subsection{Classifier-free Guidance on SDDLM}
Classifier-Free Guidance (CFG)~\cite{ho2020denoising} is an effective and versatile technique widely used in both continuous and discrete diffusion models~\cite{ho2022classifier, lovelace2024diffusion}. Grounded in Bayes’ rule, CFG jointly trains conditional and unconditional diffusion models and introduces a rescaled distribution during inference to guide generation. However, prior methods require the conditional model to be trained on paired data (prompt–response pairs) before applying classifier-free guidance (CFG). In practice, during language model pretraining, it is difficult to obtain sufficient amounts of such conditional data. To address this limitation, we adopt unsupervised CFG~\cite{nie2024scaling} and use randomly sampled tokens as the conditioning input for the unconditional generation branch:
\begin{equation}
    \tilde{p}_{\boldsymbol{\theta}}\left(\boldsymbol{x}_0 \mid \boldsymbol{c}, \boldsymbol{x}_t\right) \propto \frac{p_{\boldsymbol{\theta}}\left(\boldsymbol{x}_0 \mid \boldsymbol{c}, \boldsymbol{x}_t\right)^{1+w}}{p_{\boldsymbol{\theta}}\left(\boldsymbol{x}_0 \mid \boldsymbol{r}, \boldsymbol{x}_t\right)^w},
\end{equation}
where $\mathbf{c}$ denotes the prompt for conditional text generation, and $\mathbf{r}$ is a sequence of tokens randomly sampled from the uniform distribution over the vocabulary $\mathcal{V}$. $\tilde{p}_\theta(\cdot)$ denotes the modified distribution that replaces the original $p_\theta$ for likelihood computation and generation in Equations~(\ref{eq:usdm}) and~(\ref{eq:reverse}). The results of the CFG methods are reported in Table~\ref{tab:compare_large_cfg}, where we observe that CFG effectively improves the performance of SDDLM on different datasets.

\section{Experiment}
\subsection{Experimental Setup}
\textbf{Datasets and Models.} We evaluate our proposed method, SDDLM, along with its negative-gradient variants, SDDLM-V1, on standard language modeling benchmarks: LM1B~\cite{chelba2013one} and OpenWebText (OWT)~\cite{openwebtext}. All small models (170M-parameter) are trained for 1M steps for LM1B and OpenWebText with a batch size of 512. For LM1B, we adopt a context length of 128, while for OWT we use a context length of 1024. Moreover, to further validate the effectiveness of our method, we scale up our models to a larger parameter regime (1.1B parameters) and evaluate them on larger-scale datasets. Specifically, we employ the open-source SlimPajama dataset~\cite{shen2023slimpajama}, a multi-corpus collection containing 627 billion tokens, which is sufficiently large for all of our experiments. For fairness, we use the Llama-2 tokenizer~\cite{touvron2023llama} across all models and baselines, and set the context length to 2048.

\noindent \textbf{Implementation Details.} For small models, we follow the implementation of Duo~\cite{sahoo2025diffusion}, including the time-scheduling strategy proposed in their model. Similar to Duo, our architecture is a 170M-parameter modified Diffusion Transformer (DiT~\cite{peebles2023scalable}) with rotary positional encodings~\cite{su2024roformer} and adaptive layer normalization for conditioning on diffusion time, consistent with prior work. Training is performed on 8×H800 GPUs using bfloat16 precision. For large models, following the implementation of~\citet{nie2024scaling}, we adopt Pre-LayerNorm with RMSNorm~\cite{zhang2019root} for better stability, use SwiGLU~\cite{shazeer2020glu} as the activation function to enhance non-linearity, and implement RoPE~\cite{su2024roformer} for more expressive positional encoding. Training of large models is performed on 16×H800 GPUs using bfloat16 precision. In the following section, we use the state-of-the-art uniform state diffusion model Duo~\cite{sahoo2025diffusion} as our baseline. We denote our loss with the negative gradient defined in Equation~(\ref{eq:negative}), as SDDLM-V1. Moreover, in addition to random sampling, we also use the noisy version $\mathbf{x}_t$ itself as negative samples by denoting it as SDDLM-V2.

\noindent \textbf{Evaluation.} For small models, our zero-shot datasets include the validation splits of WikiText~\cite{merity2016pointer}, Lambada~\cite{paperno2016lambada}, AG News~\cite{zhang2015character}, and Scientific papers from ArXiv and Pubmed~\cite{cohan2018discourse}. For large models, to provide a comprehensive evaluation for large models, we assess SDDLMs on eight widely used benchmarks in the zero-shot setting, covering tasks in commonsense reasoning and reading comprehension: Hellaswag~\cite{zellers2019hellaswag}, ARC-e~\cite{clark2018think}, BoolQ~\cite{clark2019boolq}, PIQA~\cite{bisk2020piqa}, SIQA~\cite{sap2019socialiqa}, Obqa~\cite{mihaylov2018can}, RACE~\cite{lai2017race}, and LAMBADA~\cite{paperno2016lambada}. Details of datasets are put into Appendix~\ref{app:datasets}.

\noindent \textbf{Baseline.} The primary baselines for our method are the state-of-the-art Duo~\cite{sahoo2025diffusion}, USDMs (SEDD Uniform~\cite{lou2023discrete}, denoted as SEDD for simplicity, and UDLM~\cite{schiff2024simple}) and Gaussian diffusion method, PLAID~\cite{gulrajani2023likelihood}. For a fair comparison, we directly adopt the implementation with the best hyperparameters and reported results from the original paper. The details of all baselines are put into Appendix~\ref{app:baseline}.

\subsection{Sample Quality Comparison}

\begin{table}[t]
\centering
\small
\renewcommand{\arraystretch}{1.1}
\setlength{\tabcolsep}{5pt}
\begin{tabular}{l|cc|cc}
\toprule
\multirow{2}{*}{\textbf{Model}} &
\multicolumn{2}{c|}{\textbf{LM1B}} &
\multicolumn{2}{c}{\textbf{OWT}} \\
\cmidrule(lr){2-3} \cmidrule(lr){4-5}
 & Gen PPL ↓ & Entropy ↑ 
 & Gen PPL ↓ & Entropy ↑  \\
\midrule
% \textit{Autoregressive} \\
% Transformer$^{\ddagger}$ 
%  & 21.5 & 3.24 & 22.3
%  & 17.0 & 2.98 & 17.5 \\
% \midrule
% \textit{Diffusion (Absorbing State)} \\
% BERT-Mouth$^{*}$~\cite{wang2020bertmouth}
%  & -- & -- & 142.9
%  & -- & -- & -- \\
% D3PM Absorb~\cite{austin2021structured}
%  & -- & -- & 76.9
%  & -- & -- & -- \\
% SEDD Absorb$^{\ddagger}$~\cite{lou2023sedd}
%  & 33.1 & 3.72 & 32.7
%  & 24.3 & 3.41 & 24.1 \\
% MDLM~\cite{sahoo2024mdlm}
%  & 27.2 & 3.51 & 27.0
%  & 23.6 & 3.29 & 23.2 \\
% \midrule
% \textit{Diffusion (Uniform-State / Gaussian)} \\
% D3PM Uniform~\cite{austin2021structured}
%  & -- & -- & 137.9
%  & -- & -- & -- \\
% UDLM$^{\ddagger}$~\cite{schiff2025udlm}
%  & 31.3 & 3.56 & 31.3
%  & 27.4 & 3.33 & 27.4 \\
% \rowcolor{gray!10}
Duo
 & 172.93 & \textbf{4.20} & 80.43 & \textbf{5.55} \\ % & 178.44 & 4.21 & 109.93 & 5.38  \\
 SDDLM
 & 173.04 & \textbf{4.20} & 77.07 & 5.53 \\ % & 177.85 & \textbf{4.22} & 108.24 & 5.59  \\
  SDDLM-V1
 & 116.84 & 4.10 & \textbf{45.18} & 5.31 \\ % & 120.44 & 4.09 & 83.04 & 5.43 \\
  SDDLM-V2
 & \textbf{101.32} & 4.12 & 50.05 & 5.33 \\ % & \textbf{98.26} & 4.12 & 55.34 & 5.40 \\
\bottomrule
\end{tabular}
% \vskip -1em
\caption{Comparison of Gen PPL and Entropy across LM1B and OWT on SDDLM, SDDLM-V1, SDDLM-V2 and Duo.}
\label{tab:main_results_metrics}
\end{table}

\begin{figure}
    \centering
    \includegraphics[width=0.96\linewidth]{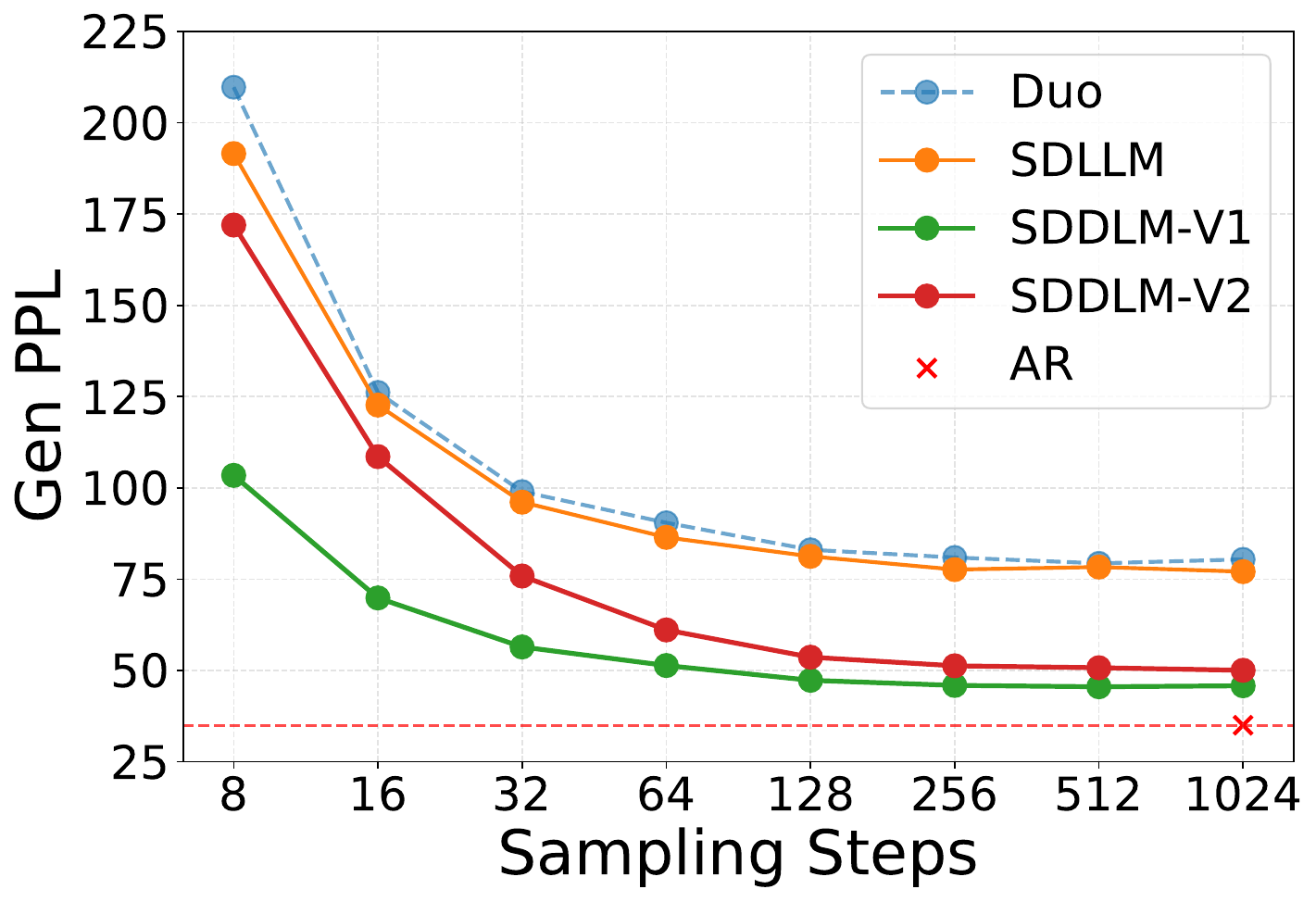}
    \vskip -0.8em
    \caption{Sample quality comparison of SDDLM and its variant vs. Duo. SDDLM-V1 outperforms Duo in Gen PPL ($\downarrow$) for base models across different sampling steps.}
    \label{fig:steps}
\end{figure}

\begin{figure*}[t]
    \centering
    \includegraphics[width=0.99\linewidth]{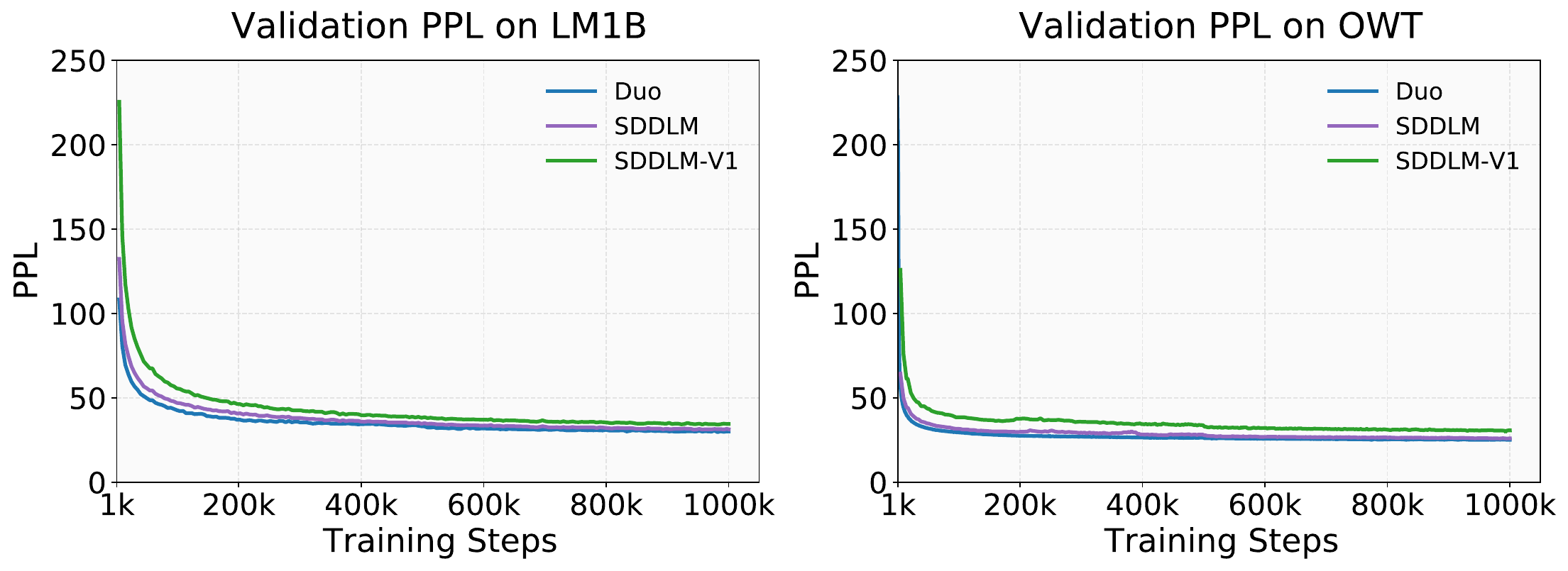}
    \vskip -1em
    \caption{Validation PPL of different models over training steps.}
    \label{fig:eppl_curve_lm1b}
\end{figure*}

\begin{figure*}[t]
    \centering
    \includegraphics[width=0.96\linewidth]{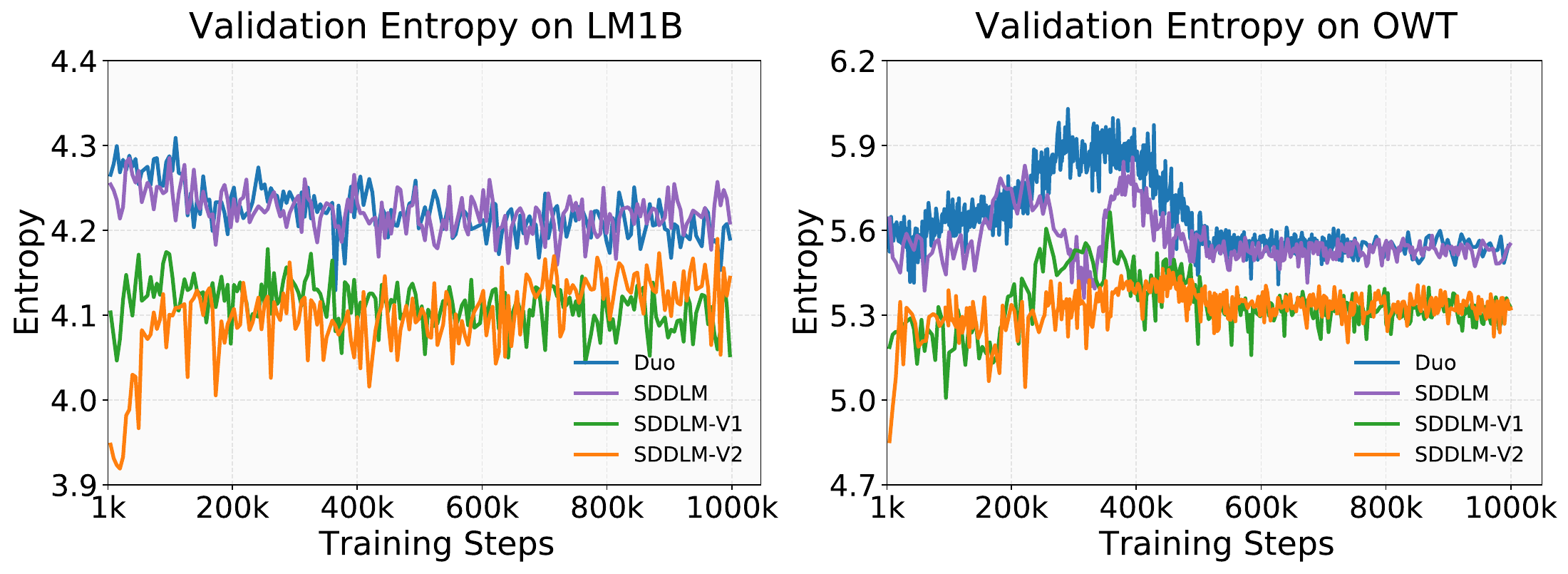}
    \vskip -1em
    \caption{Validation entropy of different models over training steps.}
    \label{fig:entropy_curve_lm1b}
\end{figure*}

To evaluate sample quality, we report GPT-2 Large generative perplexity (Gen PPL) as a measure of fluency and average sequence entropy as an indicator of diversity. The corresponding results are presented in Figure~\ref{fig:ppl_curve_lm1b} and~\ref{fig:entropy_curve_lm1b} on LM1B for validation over different training steps. Specifically, we first sample a small subset to validate the Gen PPL and entropy metrics shown in these figures, using 1,000 sampling steps. We then report the final results based on a larger set of samples generated from the final checkpoint with 1,024 sampling steps, as presented in Table~\ref{tab:main_results_metrics}. We observe that our proposed denoising loss (Equation (\ref{eq:denoising})) achieves performance comparable to the baseline in terms of both Gen PPL and entropy. Furthermore, incorporating the negative gradient loss significantly improves Gen PPL, indicating stronger alignment with real-world generative quality. These findings suggest that this simple adaptation is effective for generation quality. In Section~\ref{sec:condlanguage}, we also further explore the role of negative gradients in enhancing performance on larger models and more complex tasks. 

In addition, we evaluate our method under different numbers of sampling steps, as shown in Figure~\ref{fig:steps}. We observe that SDDLM, despite its high training efficiency, achieves performance comparable to approaches based on more complex ELBO-style objectives such as state-of-the-art baselines (Duo) across a range of sampling steps. Furthermore, the proposed regularization term consistently improves generation quality at different sampling steps. These results further demonstrate the strong generalization of our simple training objective and its enhanced variant with additional regularization across different numbers of sampling steps. Moreover, the negative-gradient–based methods show promise for reducing the required number of sampling steps (achieve better performance than Duo while using fewer sampling steps), thereby enabling the generation of more tokens within the same computational budget and improving the overall efficiency of USDMs.

Finally, we report the training cost on the OWT dataset in Figure~\ref{fig:time}. Our results show that incorporating the negative gradient leads to both improved performance and reduced training cost, measured in terms of wall-clock training time and GPU memory consumption. In addition, despite its simplicity, SDDLM with our proposed loss achieves performance comparable to Duo trained with the ELBO objective. Taken together with the previous observations, these results demonstrate that our method provides a promising and scalable approach for training USDMs, enabling efficient large-scale training with a simple yet effective loss function.

\begin{table}[t!]
\centering
{\footnotesize
\resizebox{\linewidth}{!}{
\begin{tabular}{llcccccc}
\toprule
&  Wikitext & Lambada  & AG News & Pubmed & Arxiv\\
\midrule
% \multicolumn{7}{l}{\textit{Autoregressive}} \\
% & Transformer$^\dagger$  & 25.75 & 51.28 & {52.09} & 49.01 & 41.73\\
% \midrule
% \multicolumn{7}{l}{\textit{Diffusion (absorbing state)}} \\
% & SEDD Absorb$^\dagger$ &  34.28 & 49.86 & 62.09 & 44.53 & 38.48 \\
% & D3PM Absorb$^\dagger$ &  50.86 & 93.47 & - & - & - \\
% & MDLM$^\dagger$  &  {32.83}  & 47.52 & {61.15} & 41.89 & 37.37 \\
% \midrule
% \multicolumn{7}{l}{\textit{Diffusion (Uniform-state / Gaussian)}} \\
  SEDD $^\dagger$ &    41.10  & 57.29 &  82.64 &  55.89 &  50.86 \\
  Plaid$^\dagger$ &   50.86 &  57.28 &  - &  - & - \\
  UDLM$^\dagger$ &  39.42  &  53.57 &  80.96 &  50.98 &   44.08 \\
  Duo$^\ddag$  &   \textbf{33.91}  &   \textbf{51.29} &  \textbf{69.71} &   \underline{45.34} &   \underline{40.41} \\
\midrule
   \textbf{SDDLM}  &    \underline{35.02} &    \underline{51.50} &  \underline{73.44} &   \textbf{44.75} &   \textbf{39.95}\\

\bottomrule
\end{tabular}
}
}
% \vskip -0.8em
\caption{Zero-shot perplexities ($\downarrow$) of models trained for 1M steps on OWT datasets.
All perplexities for diffusion models are upper bounds. $^\dagger$ Taken from~\citet{sahoo2025diffusion}. Best uniform / Gaussian diffusion values are \textbf{bolded} and second best values are \underline{underlined}. $^\ddag$ denotes retrained model.
}
\vspace*{-3mm}
\label{tab:zeroshot-ppl}
\end{table}

\subsection{Likelihood Evaluation}

\begin{table*}[t!]
    \centering
    
    \begin{adjustbox}{max width=\textwidth}
    \begin{tabular}{lccccccccc}
      \toprule
         & ARC-e & BoolQ & Hellaswag &  Obqa &  PIQA & RACE & SIQA  & LAMBADA \\
        \midrule
        Duo & 42.85 & \textbf{60.55} & 34.47 & 17.00 & 63.11 & 31.39 & 37.82 & 17.02 \\
        SDDLM & 45.08 & 60.34 & 34.60 & \textbf{18.20} & \textbf{64.42} & 30.05 & 37.56 & 26.53 \\
        SDDLM-V1 & \textbf{45.29} & 59.66 & \textbf{34.81} & 18.00 & 64.36 & \textbf{31.58} & \textbf{38.18} & \textbf{31.75} \\
      \bottomrule
    \end{tabular}
    \end{adjustbox}
    % \vskip -0.6em
    \caption{\textbf{Evaluation of our larger (1.1B) models.} We scale up our model and evaluate it following the methodology described in Section~\ref{sec:imple}. We assess its performance on a diverse set of QA tasks as well as next-word prediction to comprehensively evaluate the capabilities of the larger models. Large values represent better performance in the table.}
    \label{tab:compare_large}
\end{table*}

In this section, we estimate the negative log-likelihood using the ELBO, following the formulation in Duo Models~\cite{sahoo2025diffusion}. The corresponding results across different training steps are illustrated in Figure~\ref{fig:eppl_curve_lm1b}. We observe that our proposed denoising loss (Equation (\ref{eq:denoising}), SDDLM) and negative gradient with random sampling (SDDLM-V1) lead to a slight increase in perplexity (PPL) when evaluated under the ELBO framework. It is normal that our model does not directly optimize the likelihood, as similar observations have been reported in previous works on masked diffusion language models~\cite{deschenaux2024beyond}. Moreover, this discrepancy is not necessarily correlated with generation quality — in fact, we observe a substantial improvement in the quality of generated samples. Interestingly, when applying negative sampling on the perturbed sequence $\mathbf{x}_t$ (SDDLM-V2), the ELBO-based PPL decreases substantially, yet the sampling quality improves significantly. This intriguing phenomenon suggests that optimizing purely for ELBO-based likelihood may not fully capture generation quality. We further investigate this behavior on larger models and more complex tasks in Section~\ref{sec:condlanguage}. 

Moreover, we measure the zero-shot generalization of the models trained on OWT by evaluating their PPL on 5 other datasets. The corresponding results are shown in Table~\ref{tab:zeroshot-ppl}. We observe that SDDLM, which exhibits high training efficiency, achieves exact perplexity (PPL) performance comparable to the state-of-the-art method Duo, while outperforming other baseline approaches. Note that our model does not directly optimize the upper bound of the log-likelihood, whereas Duo explicitly optimizes this objective, which closely matches the evaluation protocol used in Table~\ref{tab:zeroshot-ppl} (i.e., using an upper bound rather than the true log-likelihood). Therefore, it is expected that our model does not outperform methods that directly optimize the evaluation objective, and we observe a slight drop in the metric, consistent with similar observations reported in~\citet{deschenaux2024beyond}, when optimizing a loss that differs from the evaluation metric. Importantly, this minor decrease does not affect performance on conditional language generation (Details are in Section~\ref{sec:condlanguage}), which is more commonly used as the final evaluation criterion for larger models. This may be because diffusion language models are typically evaluated using an upper bound on the log-likelihood rather than the true log-likelihood; consequently, even when the upper bound exhibits a slight decrease, the models may still achieve comparable performance on the true likelihood.

\subsection{Conditional Language Generation}

% \begin{table*}[t!]
%     \centering
    
%     \begin{adjustbox}{max width=\textwidth}
%     \begin{tabular}{lccccccccc}
%       \toprule
%          & ARC-e & BoolQ & Hellaswag &  Obqa &  PIQA & RACE & SIQA  & LAMBADA \\
%         \midrule
%        GPT-2 (1.5B) & 51.05 & 61.77 & 50.89 & 32.00 & \textbf{70.51} & 33.11 & 40.28 & 44.61 \\
%        TinyLlama (1.1B) &  \textbf{52.19} & 59.39 & \textbf{54.07} & 33.20 & 70.29 & \textbf{35.60} & 39.41 & 43.22 \\
%         MDM (1.1B) & 48.74 & \textbf{62.17} & 51.83 & \textbf{33.40} & 69.53 & \textbf{35.60} & \textbf{41.04} & \textbf{52.73} \\
%         \midrule
%         Llama-2 (7B) & 74.49 & 77.68 &75.98& 44.20 & 79.00 & 39.52 & 46.11 &68.00 \\
%       \bottomrule
%     \end{tabular}
%     \end{adjustbox}
%     \caption{\textbf{Evaluation of our 1.1B models.} Both the MDM and Llama-2 models are fine-tuned for GSM8K, with all other benchmarks assessed in zero-shot settings. Result marked $^*$ is from~\cite{gong2024scaling}. The pre-training datasets consist of approximately $540$B tokens for TinyLlama and MDM, compared to $2$T tokens for Llama-2. Our 1.1B MDM outperforms the same-size TinyLlama on four out of eight tasks and surpasses the larger GPT-2 on six tasks. Notably, MDM achieves GSM8K accuracy comparable to that of Llama-2, requiring less than $5\%$ of its pre-training FLOPs.}
%     \label{tab:compare_mdm_gpt2}
% \end{table*}

\label{sec:condlanguage}

In this section, we investigate the performance of our proposed method on conditional generation with larger models, a core language modeling task that has remained largely unexplored for USDMs. The corresponding results are reported in Table~\ref{tab:compare_large}. We find that, despite its lower computational cost and favorable scaling properties, our method achieves conditional generation performance comparable to approaches based on more complex ELBO-style objectives, such as Duo. Moreover, SDDLM-V1, equipped with an anti-uniform distribution sharpening regularization, attains performance on par with prior methods while demonstrating notably stronger results on next-word prediction tasks (e.g., Lambada). These empirical findings validate the effectiveness of the proposed regularization in mitigating the uniform distribution corruption issue. Overall, this simple yet effective regularization offers a promising direction for scaling USDMs to larger models with improved performance. More results about CFG are put into Appendix~\ref{add:exp}.

\section{Conclusion}

We have shown how to simplify the training objective of Uniform State Diffusion Models (USDMs). Inspired by the denoising loss used in continuous diffusion models, we show that this minimal objective achieves performance comparable to prior, more complex formulations. Building on this observation, we introduce a negative-gradient mechanism motivated by a self-supervised learning perspective, which further improves the model. Notably, although the estimated perplexity (PPL) derived from the ELBO increases, the  quality of the generated samples improves substantially, highlighting a mismatch between ELBO-based metrics and generation performance in USDMs. Finally, we demonstrate that the proposed approach scales effectively to larger models, with consistent empirical trends, underscoring its potential for efficient large-scale training in the future.

%In this paper, we investigate how to simplify the training objective of Uniform State Diffusion Models (USDMs). Inspired by the denoising loss used in continuous diffusion models, we show that this simple objective can achieve performance comparable to prior, more complex formulations. Building on this insight, we introduce a negative-gradient mechanism that enhances learning from a self-supervised perspective. Interestingly, although the estimated perplexity (PPL) derived from the ELBO increases, overall generation quality improves substantially. Finally, we demonstrate that our method scales effectively to larger models, where our conclusions continue to hold, highlighting the strong potential of our approach for efficient large-scale training.

\newpage
\section*{Impact Statement}
The primary goal of this work is to advance  Machine Learning. The main societal consequences of this work are likely to stem from improved performance and algorithmic efficiency gains, as is the case with most fundamental research on machine learning.

% \bibliography{ref}
% \bibliographystyle{plain}

% \appendix

% In the unusual situation where you want a paper to appear in the
% references without citing it in the main text, use \nocite
% \nocite{langley00}

\bibliography{example_paper}
\bibliographystyle{icml2025}

%%%%%%%%%%%%%%%%%%%%%%%%%%%%%%%%%%%%%%%%%%%%%%%%%%%%%%%%%%%%%%%%%%%%%%%%%%%%%%%
%%%%%%%%%%%%%%%%%%%%%%%%%%%%%%%%%%%%%%%%%%%%%%%%%%%%%%%%%%%%%%%%%%%%%%%%%%%%%%%
% APPENDIX
%%%%%%%%%%%%%%%%%%%%%%%%%%%%%%%%%%%%%%%%%%%%%%%%%%%%%%%%%%%%%%%%%%%%%%%%%%%%%%%
%%%%%%%%%%%%%%%%%%%%%%%%%%%%%%%%%%%%%%%%%%%%%%%%%%%%%%%%%%%%%%%%%%%%%%%%%%%%%%%
\newpage
\appendix
\onecolumn

\section{ Omitted Details of Derivations}

\subsection{Derivation of Equation (\ref{eq:negative})}
\label{sec:dervinegative}
Firstly, the regularization term in Equation (\ref{eq:plusregular}) can be written in the following term:
\begin{equation}
    \operatorname{KL}\left(\mathcal{U} \| p_\theta\left(\cdot \mid x_t\right)\right) = - \mathbb{E}_{\hat{x} \sim \mathcal{U}(V)} \log p_\theta\left(\hat{x} \mid x_t\right) - \log |\mathcal{V}|.
\end{equation}
Based on this Equation, we can the final objective for Equation (\ref{eq:negative}) by ignoring $\log|\mathcal{V}|$ without gradient:
\begin{equation}
    \mathcal{L}_\text{SDDLM-V1} =  \mathbb{E}_{\mathbf{x}_0, t, \mathbf{x}_t \sim \mathcal{D},  \mathcal{U}, q_t} \sum_{l=1}^L (-\log p_\theta (\mathbf{x}_0^l  \mid \mathbf{x}_t)  + \mathbb{E}_{\hat{\mathbf{x}}^l \sim \operatorname{U}(\mathcal{V})} \log p_\theta (\hat{\mathbf{x}}^l \mid \mathbf{x}_t))\mathbf{1}\left[\mathbf{x}_0^l\neq\mathbf{x}^l_t\right].
\end{equation}

\section{Details of Hyperparameter}

In this section, we put details of hyperparameters of training for Duo, SDDLM, SDDLM-V1, SDDLM-V2 in Table~\ref{tab:training_hyper}. We follow the settings used in Duo for fair comparison. Moreover, we set  $w=1.1$ for training and evaluation separately.
\begin{table}[htbp]
\caption{Training Hyperparameters for small models (170M).}
\label{tab:training_hyper}
\centering
\begin{tabular}{lllll}
\toprule
\textbf{Hyperparameter} & \textbf{Duo} & \textbf{SDDLM} & \textbf{SDDLM-V1} & \textbf{SDDLM-V2} \\
\midrule
Optimizer & AdamW & AdamW & AdamW & AdamW \\
Learning Rate & $3\times10^{-4}$ & $3\times10^{-4}$ & $3\times10^{-4}$ & $3\times10^{-4}$ \\
LR Schedule & Linear Decay & Linear Decay & Linear Decay & Linear Decay \\
Warm-up Steps & 2500 & 2500 & 2500 & 2500 \\
Decay Rate $\beta_1$ & 0.9 & 0.9 & 0.9 & 0.9 \\
Decay Rate $\beta_2$ & 0.999 & 0.999 & 0.999 & 0.999 \\
Weight Decay & 0 & 0 & 0 & 0 \\
Global Batch Size & 512 & 512 & 512 & 512 \\
Training Steps & 1000k & 1000k & 1000k & 1000k \\
EMA Decay & 0.9999 & 0.9999 & 0.9999 & 0.9999 \\
\bottomrule
\end{tabular}
\end{table}

\begin{table}[htbp]
\caption{Training Hyperparameters for Large models (1.1B).}
\label{tab:training_hyper_large}
\centering
\begin{tabular}{lllll}
\toprule
\textbf{Hyperparameter} & \textbf{Duo} & \textbf{SDDLM} & \textbf{SDDLM-V1}  \\
\midrule
Optimizer & AdamW & AdamW & AdamW \\
Learning Rate & $2\times10^{-4}$ & $2\times10^{-4}$ & $2\times10^{-4}$ \\
LR Schedule & Linear Decay & Linear Decay & Linear Decay \\
Decay Rate $\beta_1$ & 0.9 & 0.9 & 0.9 \\
Decay Rate $\beta_2$ & 0.95 & 0.95 & 0.95 \\
Weight Decay & 0.1 & 0.1 & 0.1  \\
Global Batch Size & 256 & 256 & 256  \\
\bottomrule
\end{tabular}
\end{table}
\section{Additional Details of Experiments}

\subsection{Details of Datasets}
\label{app:datasets}
In this section, we provide an overview of the benchmarks used in the Experiment section.

\textbf{WikiText}~\cite{merity2016pointer}. This dataset consists of high-quality Wikipedia articles and is commonly used to assess general language modeling performance. 

\textbf{Lambada}.~\cite{paperno2016lambada} This dataset focuses on long-range dependency modeling, requiring the prediction of the final word in a passage given a broad context. 

\textbf{AG News}.~\cite{zhang2015character} This dataset is a news corpus covering multiple topic categories and evaluates models’ ability to model news-style text. 

\textbf{ArXiv and PubMed}.~\cite{cohan2018discourse} These datasets contain scientific articles from their respective domains and are used to benchmark language modeling performance on long, technical documents.

For AG News and Scientific Papers (PubMed and ArXiv), we apply both the WikiText and One Billion Words detokenizers. Since the zero-shot datasets follow different conventions for sequence segmentation, we wrap all sequences to a length of 1024 tokens and do not insert end-of-sequence (EOS) tokens between consecutive segments.

\textbf{ARC-Easy.}~\cite{clark2018think} A subset of the AI2 Reasoning Challenge that concentrates on elementary-level science questions, designed to evaluate a model’s reasoning ability based on fundamental scientific concepts.

\textbf{BoolQ.}~\cite{clark2019boolq} A yes-or-no question-answering dataset designed to evaluate a model’s ability to answer questions based on a given passage.

\textbf{HellaSwag.}~\cite{zellers2019hellaswag} A dataset assesses the model’s commonsense reasoning ability by completing a given sentence with one of four options.

\textbf{OpenBookQA.}~\cite{mihaylov2018can} A question-answering dataset modeled after open-book exams, designed to assess a model’s understanding of a subject by requiring multi-step reasoning and the integration of additional commonsense knowledge.

\textbf{PIQA.}~\cite{bisk2020piqa} Physical Interaction Question Answering is a dataset that evaluates physical reasoning ability by asking models to select the best solution to problems involving everyday physical scenarios..

\textbf{SIQA.}~\cite{sap2019socialiqa} Social Interaction Question Answering is a commonsense reasoning benchmark that presents scenarios requiring models to reason about social interactions and the motivations underlying human behavior.

\textbf{RACE.}~\cite{lai2017race} ReAding Comprehension Dataset From Examinations was designed to evaluate reading comprehension ability through understanding and interpreting high school–level texts.

\textbf{LAMBADA.}~\cite{paperno2016lambada} A dataset designed to evaluate models’ text understanding capabilities through a final single-word prediction task based on a given context. 

\subsection{Details of Baselines}
\label{app:baseline}
In this section, we provide detailed descriptions of baselines in experiments.

\textbf{Duo}.~\cite{sahoo2025diffusion} This method improves USDMs by leveraging their connection to Gaussian diffusion, introducing Gaussian-guided curriculum learning for efficient training and discrete consistency distillation for fast few-step generation.

\textbf{SEDD}.~\cite{lou2023discrete} It introduces score entropy to extend score matching to discrete domains, enabling effective diffusion-based language modeling with strong perplexity and generation quality compared to both prior diffusion and autoregressive models.

\textbf{UDLM}.~\cite{schiff2024simple} This work extends classifier-based and classifier-free guidance to discrete diffusion models and introduces uniform-noise diffusion with a continuous-time training objective, enabling strong controllable generation across diverse discrete domains.

\textbf{PLAID}.~\cite{gulrajani2023likelihood} This work improves the likelihood performance of diffusion language models via algorithmic advances and scaling law analysis.

\subsection{Additional Experiments}
\label{add:exp}
In this section, we present additional results for CFG applied to different algorithms, as shown in Table~\ref{tab-add:compare_large_cfg}. We observe that CFG consistently improves performance across these methods. In particular, our simple loss formulation achieves comparable performance on QA tasks and superior results on next-word prediction. Moreover, incorporating our proposed negative-gradient regularization further enhances performance on next-word prediction tasks.

\begin{table*}[t!]
    \centering
    
    \begin{adjustbox}{max width=\textwidth}
    \begin{tabular}{lccccccccc}
      \toprule
         &  BoolQ & Hellaswag &  Obqa &  PIQA & RACE & SIQA  & LAMBADA \\
        \midrule
        % Duo & 42.85 & \textbf{60.55} & 34.47 & 17.00 & 63.11 & 31.39 & 37.82 & 17.02 \\
        Duo-CFG &  \textbf{62.20} & 36.00 & 18.80 & 63.22 & \textbf{31.67} & 37.10 & 20.53 \\
        SDDLM-CFG &  62.11 & 36.09 & \textbf{20.00} & \textbf{64.80} & 30.72 & 37.87 & 32.60 \\
        SDDLM-V1-CFG & 62.17 & \textbf{36.22} & 19.00 & 64.47 & 29.86 & \textbf{38.95} & \textbf{34.70} \\
      \bottomrule
    \end{tabular}
    \end{adjustbox}
    % \vskip -0.7em
    \caption{Additional Results for CFG methods on 1.1B models.}
    \label{tab-add:compare_large_cfg}
\end{table*}

\subsection{Details of Training and Evaluation}
\label{sec:imple}
\begin{algorithm}[tb]
   \caption{SDDLM \& SDDLM-V1}
   \label{alg:training}
\begin{algorithmic}
    \STATE {\bfseries Input:} Epochs $N$, trainable parameters $\theta$ 
    \FOR{$i=1$ {\bfseries to} $N$}
        \STATE $\mathbf{x}_0 \sim q(\mathbf{x}_0)$
        \STATE $t \sim \mathrm{Uniform}[0,1]$
        \STATE $\mathbf{x}_t \sim q_t\left(. \mid \mathbf{x}^l_0 ; \alpha_t\right)$, $\hat{\mathbf{x}}^l \sim \mathcal{U}$ (for SDDLM-V1)
        \STATE Take gradient descent step on
        \STATE  $\nabla_\theta \mathcal{L}_{\text{SDDLM}}$ or $\nabla_\theta \mathcal{L}_{\text{SDDLM-V1}}$
    \ENDFOR
    \STATE \textbf{return} $\theta$
\end{algorithmic}
\end{algorithm}

\textbf{Training}. Both algorithms are summarized in Algorithm~\ref{alg:training}. They are deliberately simple: training proceeds by sampling either positive examples only, or both positive and negative examples (for SDDLM-V1), and optimizing the model under the corresponding objectives. Moreover, we adopt curriculum learning~\cite{bengio2009curriculum}, in which the model is trained on increasingly complex data, starting from simpler, easier-to-denoise noise patterns and gradually progressing to more challenging ones. Following \citet{sahoo2025diffusion}, this strategy has been shown to be effective for USDMs, and we adopt this curriculum to accelerate training. Notably, when scaling to larger models with 1.1B parameters in our experiments, our method can be readily applied and demonstrates strong potential for efficient scaling.

\textbf{Sampling and Evaluation}. For unconditional sampling, we iteratively sample from Equation~(\ref{eq:reverse}) using the trained model, starting from a uniform distribution that randomly samples token sequences from the vocabulary. For conditional generation, we compute the model likelihood as defined in Equation~\ref{eq:usdm} and apply it to QA tasks.

\section{Generation Examples}

To ensure correct LaTeX rendering, we manually process the generated text following Duo~\cite{sahoo2025diffusion}:

\begin{enumerate}
    \item Curly double quotes \verb|(\u201c, \u201d)| replaced with "
    \item Em dashes/en dashes \verb|(\u2014, \u2013)| replaced with -- or -
    \item Soft hyphens \verb|(\u00ad)| removed (or replaced by a normal hyphen where it makes sense)
    \item Any other special characters replaced with a suitable ASCII approximation
\end{enumerate}

\SampleBox{Samples from model on OWT obtained from SDDLM. Preplexity=77.07, Entropy=5.53.}{./text/SDDLM_sample.txt}

\SampleBox{Samples from model on OWT obtained from SDDLM-V1. Preplexity=45.18, Entropy=5.31.}{./text/SDDLM_v1_sample.txt}

\SampleBox{Samples from model on OWT obtained from SDDLM-V2. Preplexity=50.05, Entropy=5.33.}{./text/SDDLM_v2_sample.txt}

\SampleBox{Samples from model on OWT obtained from Duo. Preplexity=80.43, Entropy=5.55.}{./text/duo_sample.txt}

\end{document}